\pdfoutput=1

\documentclass{article} 
\usepackage{latex_template,times}
\usepackage{hyperref}
\usepackage{url}
\usepackage{xcolor}
\usepackage{listings}
\usepackage{amssymb}
\usepackage{amsmath}
\usepackage{biblatex}
\usepackage[justification=centering]{caption}
\usepackage{mathtools}
\usepackage[many]{tcolorbox}
\usepackage{color,soul}
\usepackage{pdflscape}
\usepackage[section]{placeins}

\title{Analysis of the robustness of NMF algorithms}

\nipsfinalcopy 

\begin{document}
\maketitle

\author{}
\author{
\begin{tabular}[t]{@{\extracolsep{1em}}*3c} 
\textbf{Alejandro Díaz} & \textbf{Damian Steele} & \textbf{Dan Dinh Nguyen}\\
adia2600@uni.sydney.edu.au & dste5943@uni.sydney.edu.au & mdin6872@uni.sydney.edu.au
\end{tabular}
}
\vskip 0.4in

\begin{abstract}
We examine three non-negative matrix factorization techniques; $L_{2}$-norm, $L_1$-norm, and $L_{2,1}$-norm. Our aim is to establish the performance of these different approaches, and their robustness in real-world applications such as feature selection while managing computational complexity, sensitivity to noise and more. We thoroughly examine each approach from a theoretical perspective, and examine the performance of each using a series of experiments drawing on both the ORL and YaleB datasets. We examine the Relative Reconstruction Errors (RRE), Average Accuracy and Normalized Mutual Information (NMI) as criteria under a range of simulated noise scenarios.
\end{abstract}

\section{Introduction}
Non-negative matrix factorization (NMF) can play a role in several machine learning applications, not least feature selection but also in signals processing and error-correction. A consequence of NMF is a sparse representation of basis elements that may underlie a multi-dimensional domain, and it is this property that we can exploit for real-world applications. Our aim is to evaluate three techniques in a effort to establish what considerations a machine learning practitioner might wish to consider when faced with corrupted or otherwise noisy data using robust methods.

We baseline the perennial $L_2$-norm, and contrast its performance in simulated scenarios with two other techniques known to provide robust performance which we measure using Reconstruction Errors (RRE), Average Accuracy and Normalized Mutual Information (NMI). Based on extensive review of the literature, we've chosen to examine the robust $L_1$-norm, and robust $L_{2,1}$-norm in-depth. Both $L_1$-norm and $L_{2,1}$-norm have been shown to be less sensitive to outliers, while $L_{2,1}$-norm has also been established to be computationally efficient to iteratively approximate \cite{Nie10efficientand}.

Our results demonstrate the practical benefits of robust NMF, and lead us to believe an opportunity may exist in further exploring the sensitivity of these methods to the initial starting-state of the factor matrices.

\pagebreak

\section{Related Work}
\label{sec:basic_nmf}
The Non-negative Matrix Factorization (NMF) algorithms have become very popular for the analysis of high-dimensional data due to its ability to automatically handle sparsity problems and extract interpretable features \cite{gillis}.

Given a non-negative matrix $X \in \mathbb{R}^{mxn}$, each column of $X$ represents a data sample (in this case an image), the NMF algorithm ideally aims to find two non-negative matrices $U \in \mathbb{R}^{mxk}$ and $V \in \mathbb{R}^{kxn}$ for approximating $X$ by the product of them (Equation \ref{eq:nmf}).

\begin{equation}
    X \approx UV
\label{eq:nmf}
\end{equation}

Minimizing the following objective function we learn the matrices $U$ and $V$ \cite{alg_nmf}:

\begin{equation}
    \min_{U \in \mathcal{U},\; V \in \mathcal{V}} \|X - UV\|^2_{F}
\end{equation}

where $\|.\|_F$ indicates the Frobenius norm. Additionally, the matrices $U$ and $V$ must be non-negative:

\begin{equation}
  \begin{aligned}
    \mathcal{U} = \mathbb{R}^{dxk}_+ \\
    \mathcal{V} = \mathbb{R}^{kxn}_+
  \end{aligned}
\end{equation}

The iterative multiplicative update rule (MUR) presented below is used to minimize the above objective function:

\begin{equation}
    U_{ij} = U_{ij} \frac{(XV^T)_{ij}}{(UVV^T)_{ij}}
\end{equation}
\begin{equation}
    V_{ij} = V_{ij} \frac{(U^TX)_{ij}}{(U^TUV)_{ij}}
\end{equation}

The non-negative matrix factorization problem is non-convex, but using an iterative updating rule to calculate $U_{ij}$ while $V_{ij}$ is fixed allows us to optimize the objective function.

\section{Methods}
In this section, we provide the formulations of the different non-negative matrix factorization algorithms as well as the optimization steps. Additionally, we explain the robustness of each algorithm from a theoretical view.  

\subsection{Pre-processing}
There are numerous pre-processing techniques to be taken into consideration such as global centering/local centering, image whitening etc. we decided to use image normalization for the provided datasets. Normalization technique is widely believed to be essential not just because of its simplicity but it also helps removing noise while bringing the images into a range of intensity values that is considered "normal". This can be interpreted statistically that it follows a normal distribution as much as possible, hence its physically less stressful to our visual sense.

\subsection{L\textsubscript{1} Norm Robust Non-negative Matrix Factorization}
The L\textsubscript{1} Norm Robust NMF is particularly designed to improve robustness removing the unknown large addictive noise. It models the partial corruption, which is treated as large additive noise and it is able to simultaneously learn the basis matrix, coefficient matrix and estimate the positions and values of noise \cite{l1reg}.

\subsubsection{Loss Function}
\begin{figure}[h]
\centering
\includegraphics[width=5cm]{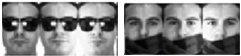}
\centering
\caption{Examples where the noise distribution is sparse. The occlusion by glasses \\ or scarves is an instance of this kind of noise.}
\label{fig:partial_noise}
\end{figure}

Let non-negative matrix $X \in \mathbb{R}^{mxn}$ denote the observed corrupted data and where each column of $X$ represents a data sample. Let $\widehat{X} \in \mathbb{R}^{mxn}$ denote the clean data without noise. The noise in the data can properly be formulated as:

\begin{equation}
    X = \widehat{X} + E
\end{equation}

where $E \in \mathbb{R}^{mxn}$ is the large additive noise. It is important to mention that the noise $E$ is not Gaussian Noise with zero mean, which is well handled by least squares error \cite{l1reg}. In this case, we focus on partial corruption, this indicates that only a small number of entries of $E$ are nonzero which it means that the noise distribution is sparse. An example of sparse noise distribution is showed in the Figure
\ref{fig:partial_noise}.

As we mentioned in the previous section \ref{sec:basic_nmf}, the clean data $\widehat{X}$ can be approximated by the equation \ref{eq:nmf}. In this case, as in traditional NMF, we can approximate it by:

\begin{equation}
    X = UV + E
\end{equation}

where $U \in \mathbb{R}^{mxk}$, $V \in \mathbb{R}^{kxn}$ and $E \in \mathbb{R}^{mxn}$. In this case, and taking into account the above model, the objective function of RobustNMF is defined by the next equation:

\begin{equation}
    O_{RobustNMF} = \|X - UV - E\|^2_F + \lambda \sum_j [\|E_{.j}\|_0]^2
    \label{eq:l1_reg_base}
\end{equation}

The clean data is approximated by the first term and the second term is obtained from the sparseness constraint of $E$. Additionally, $\lambda$ is the factor to control the ratio between both terms. As we can observe, the $L_0$ norm in the second term makes this function hard to optimize. A common technique applied is to approximate $L_0$ by $L_1$ norm \cite{l1_approx}. Rewriting the above equation with the $L_1$ norm, we have:

\begin{equation}
    \begin{aligned}
    O_{RobustNMF} = \|X - UV - E\|^2_F + \lambda \sum_j [\|E_{.j}\|_1]^2\\
     = \|X - [U,I, -I]
    \binom{V}{\begin{pmatrix} E^p\\ E^n\end{pmatrix}}\|^2_F \\
    + \lambda \sum_j [\|E^p_{.j}\|_1 + \|E^n_{.j}\|_1]^2
    \end{aligned}
    \label{eq:l1_rep}
\end{equation}

where:

\begin{equation}
    \begin{aligned}
    E = E^p - E^n\\
    E^p = \frac{|E| + E}{2}\\
    E^n = \frac{|E| - E}{2}
    \end{aligned}
\end{equation}

\label{constr}
and $E^p$ and $E^n$ are non-negative matrices thus, $E^p \geq 0$ and $E^n \geq 0$. The reason of the decomposition of $E$ into $E^p$ and $E^n$ is due to gain the non-negativity which results in the convenience in optimization \cite{l1reg}. Furthermore, we also need to set a constraint for the clean data $\widehat{X}$ since it should be non-negative thus, $X - E \geq 0$. To sum up, the objective function should be minimized with respect $U$, $V$, $E^p$ and $E^n$ with the following constraints: $U \geq 0$, $V \geq 0$, $E^p \geq 0$, $E^n \geq 0$ and $X - E \geq 0$.

\subsubsection{Optimization}
As we mentioned before, we replaced the $L_0$ norm in the equation \ref{eq:l1_reg_base} by $L_1$ norm because the $L_0$ makes this equation hard to optimize and this allowed us to simplify the problem. However, the equation $O_{RobustNMF}$ is not convex with respect $U$, $V$, $E^p$ and $E^n$, as in traditional NMF, we use a multiplicative updating algorithm to iterative updating $U$, $V$, $E^p$ and $E^n$ \cite{alg_nmf}.

The optimization of $U$, $V$ and $E$ has to be done iteratively by fixing the others. We start initializing randomly these matrices, then we update $U$ and finally, we update $V$ and $E$ at the same time. This is possible due to the $L_1$ norm in the equation.\\

\textbf{Initialization}

Like most of the iterative algorithms, we need to initialize $U$, $V$ and $E$ at the beginning. In this case, we use a normal distribution with mean 0 and variance 1. Besides, these matrices are forced to be non-negative due to the constrain mentioned in the section \ref{constr}.

An interesting question is to ask us if the converge of our algorithm depends on the initialization method but as \cite{l1_reg2} indicates, we find that our algorithm is not sensitive to the initialization method.

\textbf{Updating $U$}

We update $U$ with $V$, $E^p$ and $E^n$ given.

\begin{equation}
    \begin{aligned}
    U = \arg \min_{U \geq 0} \|X - [U,I, -I]
    \binom{V}{\begin{pmatrix} E^p\\ E^n\end{pmatrix}}\|^2_F \\
    + \lambda \sum_j [\|E^p_{.j}\|_1 + \|E^n_{.j}\|_1]^2 \\
    =  \arg \min_{U \geq 0} \|[X - E] - UV\|^2_F
    \end{aligned}
\end{equation}

Solving the previous equation:

\begin{equation}
    \begin{aligned}
    U_{ij} = U_{ij}\frac{(\widehat{X}V^T)_{ij}}{(UVV^T)_{ij}}
    \end{aligned}
\end{equation}

where $\widehat{X} = X - E$ and as $E^p$ and $E^n$ is given we can compute the value of $E$.

\textbf{Updating $V$, $E^p$ and $E^n$}

Once we updated $U$ and with $U$ known, we can update $V$, $E^p$ and $E^n$ to decrease the objective function with respect these matrices. The updating rule for $\widetilde{V}$ is:

\begin{equation}
    \begin{aligned}
    \widetilde{V}_{ij} = 
    \max(0, \widetilde{V}_{ij} - \frac{\widetilde{V}_{ij}(\widetilde{U}^T\widetilde{U}\widetilde{V})_{ij}}{(S\widetilde{V})_{ij}} + \frac{\widetilde{V}_{ij}(\widetilde{U}^T\widetilde{X})_{ij}}{(S\widetilde{V})_{ij}})
    \end{aligned}
\end{equation}

where: 

\begin{equation}
    \begin{aligned}
    \widetilde{X} = \binom{X}{0_{1xn}} \quad 
    \widetilde{U} = \binom{U, I, -I}{0_{1xk}\sqrt{\lambda}e_{1xm}\sqrt{\lambda}e_{1xm}} \quad 
    S_{ij} = |(\widetilde{U}^T\widetilde{U})_{ij}|
    \end{aligned}
\end{equation}

\subsubsection{Advantages}
In the real world applications, the data are usually corrupted with noise and the traditional NMF algorithm can not perform efficiently in these cases. For this reason, L\textsubscript{1} Norm Robust NMF provides us with a way to handle partial corruption and besides, it does not require the location of the noise in advance. It can locate and estimate the large additive noise and learn the basis matrix $U$ and $V$ at the same time.

Additionally, as \cite{l1reg} suggests, the position of the noise, computed by the matrix $E$, can be used in parallel with other denoising algorithms such as WNMF.

\subsection{L\textsubscript{2,1} Norm Robust Non-negative Matrix Factorization}
One of the unfortunate drawbacks of standard NMF is that is prone to outliers, and noise. Practical applications of NMF might be adversely impacted by this limitation, such as for the purposes of feature selection. $L_{2,1}$ norm based regularization has been established as a novel, efficient and robust means to apply to domains sensitive to outliers \cite{Nie10efficientand}. As previously discussed, the MUR (4,5) provides a method to iteratively update and approximate the non-convex $L_2$ objective function. We examine how this has been extended to address this limitation.
\begin{equation}
    \left \|X - FG \right \|^2_F = \sum_{i=1}^{n} \left \|x_i-Fg_i\right \|^2
\end{equation}
The error for each data point is calculated as a squared residual error in terms of $|x_i-Fg_i\|^2$. Outliers with large errors can dominate the objective function due to this squared operation \cite{10.1145}. The robust $L_{2,1}$ norm NMF replaces the aforementioned objective function to forego the squared residual errors.
\begin{equation}
    \left \|X - FG\right \|_{2,1} = \sum_{i=1}^{n} \sqrt{\sum_{j=1}^{p} (X-FG)^2_{j,i}} = \sum_{i=1}^{n} \left \|x_i-Fg_i\right \|
\end{equation}

\subsubsection{Loss Function}
Intuitively, outliers will have less importance as the squared residual errors has been removed. A consequence of this new formulation is that error function appears harder to solve as the previous iterative approximation technique (MUR) no longer provides a means to optimise the factor matrices $F$ and $G$.
 \begin{equation}
    \min_{F,G} \left \|X - FG \right \|_{2,1} \ s.t. \ F \geq 0, G \geq 0
\end{equation}
Kong (2011) extends the $L_2$ MUR, with the addition of a weighted matrix regularizer $D$ \cite{10.1145}. The $L_2$ MUR is extended to incorporate these weights which works to suppress outliers. The additional $D$ matrix might suggest a more complex solution, however the algorithm allows for a fast and straight forward implementation. We can see the $L_{2,1}$ norm is utilized as both an iteratively updated regularizer and in the objective function.

\begin{equation}
 \begin{aligned}
    \ D_{ii} \Leftarrow \frac{1}{\sqrt{\sum_{j=1}^{p}(X-FG)^2_{ji}}} = \frac{1}{\left \|x_i-Fg_i \right \|}\\
    \ F_{jk} \Leftarrow F_{jk} \frac{(XDG^T)_{jk}}{(FGDG^T)_{jk}}\\
    \ G_{ki} \Leftarrow G_{ki} \frac{(F^TXD)_{ki}}{(F^TFGD)_{ki}}
 \end{aligned}
\end{equation}

\subsubsection{Optimization}
The use of $L_{2,1}$ norm in the loss function requires a reasonably straight forward adaptation of the standard $L_2$ NMF updating rules. Firstly $D_{ii}$ is calculated (18) as a diagonal matrix. Elements of $D$ provide a weighted projection which serves to regularize outliers in features. The diagonal matrix allows us to incorporate the new term in the dot product of the factor matrices.

\begin{figure}[h]
\centering
\includegraphics[width=6cm]{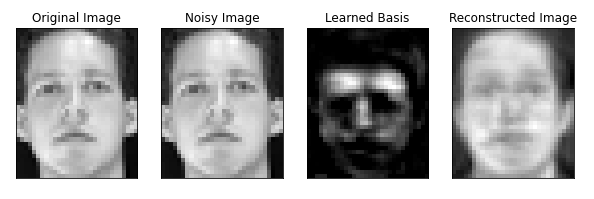}\\
\includegraphics[width=6cm]{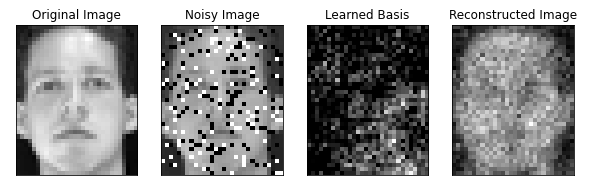}\\
\includegraphics[width=6cm]{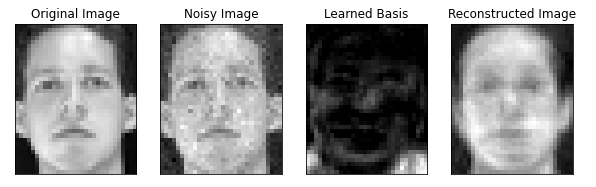}
\centering
\caption{Learned basis using $L_{2,1}$ norm NMF with no noise (Top), Salt and Pepper (Middle) (p=0.2, s vs p=0.4) and Laplace (Bottom) (loc=0, scale=6) noise.}
\label{fig:l21-verification}
\end{figure}
Initialisation of both $F_{jk}$ and $G_{kj}$ is performed using a random assignment. Updating $G$ while fixing $F$, and vice-versa, it can be shown that the Karush-Kohn-Tucker condition holds and $J(G)$ and $J(F)$ decreases monotonically \cite{10.1145}. The results of iteratively updating $D$,$F$ and $G$ can be seen in Figure \ref{fig:l21-verification} under a number of simulated noise conditions.

\subsubsection{Advantages}
The robust $L_{2,1}$ norm NMF method has several advantages, primarily being simplicity and fast performance. Practical applications of NMF, such as for feature selection, require robust and relatively stable objective functions when applied to complex domains. The theoretically more complex $L_{2,1}$ norm NMF objective function can be solved using a relatively straight forward approach.

\subsection{L\textsubscript{2} Norm MUR Non-negative Matrix Factorization}
Non-negative matrix factorization (NMF) is a widely-used tool for obtaining low-rank approximations of non-negative data such as digital images. Hence we also explore the affect of the L2 base Multiplicative Update Rules on the provided data sets. In addition, the matrices initiation is implemented using singular value decomposition based method called Non-negative Double Singular Value Decomposition (NNDSVD).

\subsubsection{Loss Function}
\begin{equation}
\min \|X - WH\|^2, w.t.W,H; s.t.W,H >0
\end{equation}

The problem is an NP-hard problem because it is convex in W or H but not in both. this problem cannot be solved analytically so it is generally approximated numerically, it means that for this minimization problem we cannot find the global minimum, however, finding a good local minimum can be satisfying for us. Therefore we are going to use an algorithm called multiplicative update rule introduced by Lee and Seung in 1999\cite{alg_nmf}.

\subsubsection{Optimization}

The algorithm is based on gradient descent with different updates rules so basically in this algorithm at first we randomly initialize W and H, in each iteration we fix the W and updates H, then we fix the H and update the W.
\begin{equation}
    H  \longleftarrow H -  \eta  \odot [W^TWH-W^TX ]
\end{equation}
\begin{equation}
    W  \longleftarrow W -   \zeta   \odot [W^TWH-W^TX ]
\end{equation}

Lee and Seung has proved that under these objects rules, gradient descent ensures the convergence of this problem in limited number of iterations. What they did is that they used additive updates rules and they set the learning rates in this way 

\begin{equation}
\eta  \longrightarrow  \frac{H}{W^TWH} 
\end{equation}

\begin{equation}
 \zeta   \longrightarrow  \frac{W}{WHH^T} 
\end{equation}

and finally they get the multiplicative update rules

\begin{equation}
H  \longleftarrow H \odot  \frac{W^TX}{W^TWH} 
\end{equation}

\begin{equation}
W  \longleftarrow W \odot  \frac{W^TX}{WHH^T} 
\end{equation}

\textbf{SVD based initialization}

As the Multiplicative Update Rule is also an iterative method, It is very sensitive to the initialization of W and H.
NNDSVD Method is a SVD based initialization, introduced by C. Boutsidis and E. Gallopoulos in 2007.\cite{boutsidis2008svd}

Non-negative Double Singular Value Decomposition (NNDSVD), non-negative matrix factorization (NMF) has been proven to be significantly enhanced by the new designed NNDSVD initiation method.
NNDSVD may be combined with almost all existing NMF algorithms. The NNDSVD initiation algorithm
does not contain randomization, additionally, it is based on approximations of positive sections of the partial singular value factors of the data matrices, using an algebraic property of unit rank matrices. Therefore, we also explored this initialization method and its effects on the randomized matrix method.

\begin{figure}[h]
\centering
\includegraphics[width=13cm]{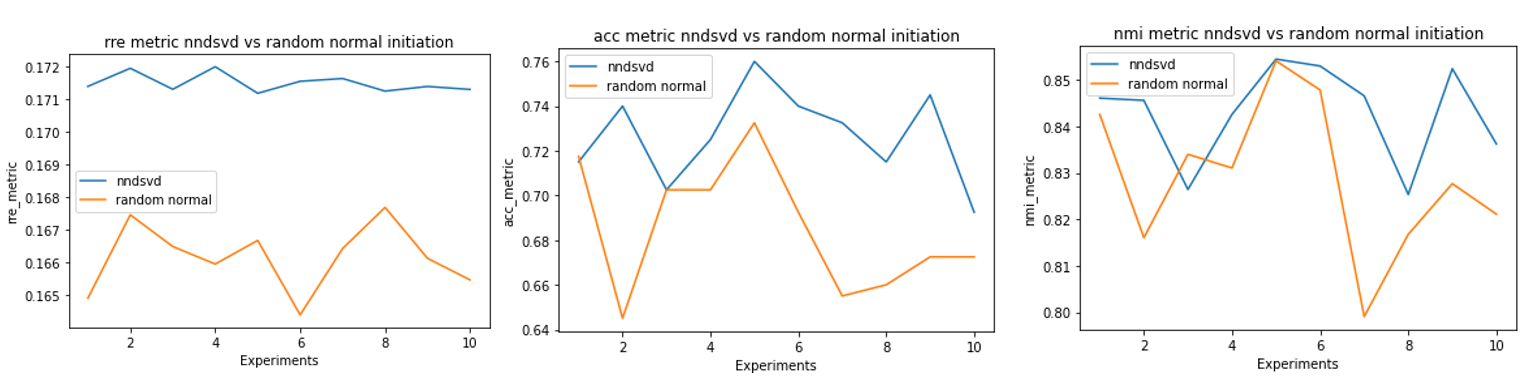}
\centering
\caption{Effects of initiation methods on evaluation metrics on ORL dataset}
\label{fig:sp_example}
\end{figure}

It is observed that applying NNSVD initiation on the matrices provided better accuracy on all 10 accuracy tests, the technique also achieved higher NMI in almost all experiments in comparison to the randomized matrices. All the tests in figure 3 was performed on ORL datasets with added Gaussian noise through 100 iterations and 30 components. 

\subsubsection{Advantages}
NNDSVD is widely believed to be the optimal initialization for NMF algorithms with sparse factors. Numerous statistical examples suggest that NNDSVD results in quicker reduction of the approximation error of many NMF implementations \cite{boutsidis2008svd}.

\subsection{Noise}
In this section, we present the formulation of the different types of noise implemented for analyzing the robustness of the Non-negative Matrix Factorization algorithms mentioned in the last section.

\subsubsection{Salt and Pepper Noise}
This is also called Impulse Valued Noise or Data Drop Noise. The Salt and Pepper Noise corrupts some pixel values in the image. The pixels with noise are replaced by corrupted pixel values where the maximum possible value is 255 (Salt noise) or 0 (Pepper noise) \cite{noise}.

\begin{figure}[h]
\centering
\includegraphics[width=8cm]{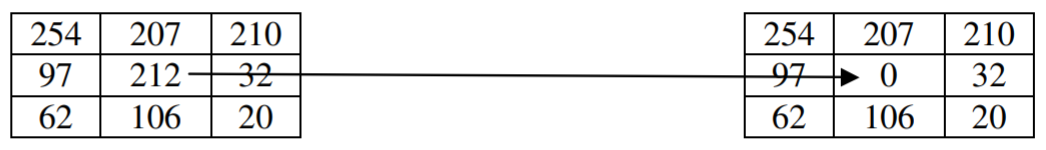}
\centering
\caption{Matrix corresponding to the pixel values in an image. The central pixel is corrupted by Pepper noise.}
\label{fig:salt_and_pepper}
\end{figure}

In the Figure \ref{fig:salt_and_pepper}, it can be observed a 3x3 matrix, corresponding to the pixel values in an image, where the central pixel is corrupted by Pepper noise.

The noise is generated by randomly changing some pixel values in the image. The parameters to tune the salt and pepper noise are \emph{p} and \emph{s\_vs\_p}. The parameter \emph{p} indicates the noise level, for instance, $p=0.1$ indicates that the $10\%$ of the pixel values in the image should be modified. On the other hand, the parameter \emph{s\_vs\_p} controls the ratio between Salt and Pepper noise. For example, if $s\_vs\_p = 0.2$, then $20\%$ of the modified pixel values should be white. The Figure \ref{fig:sp_example} shows the Salt and Pepper Noise applied to an image extracted from the ORL dataset.

\begin{figure}[h]
\centering
\includegraphics[width=8cm]{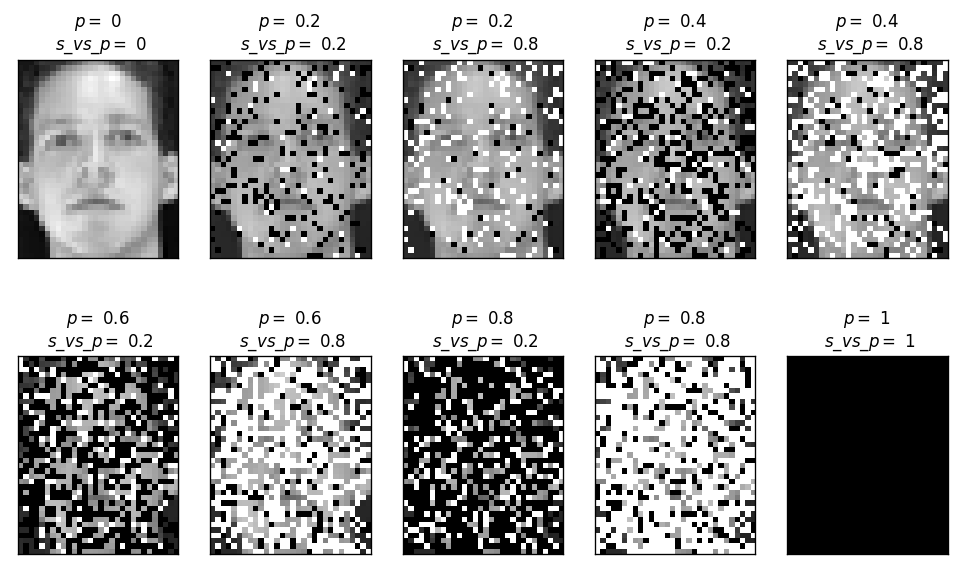}
\centering
\caption{Image with Salt and Pepper noise with different values of hyper-parameters $p$ and $s\_vs\_p$.}
\label{fig:sp_example}
\end{figure}

This noise is usually generated by malfunctioning of pixel elements in camera sensors, errors in conversion process or lack of storage space and many more \cite{noise}.

\subsubsection{Gaussian Noise}
\begin{figure}[h]
\centering
\includegraphics[width=5.5
cm]{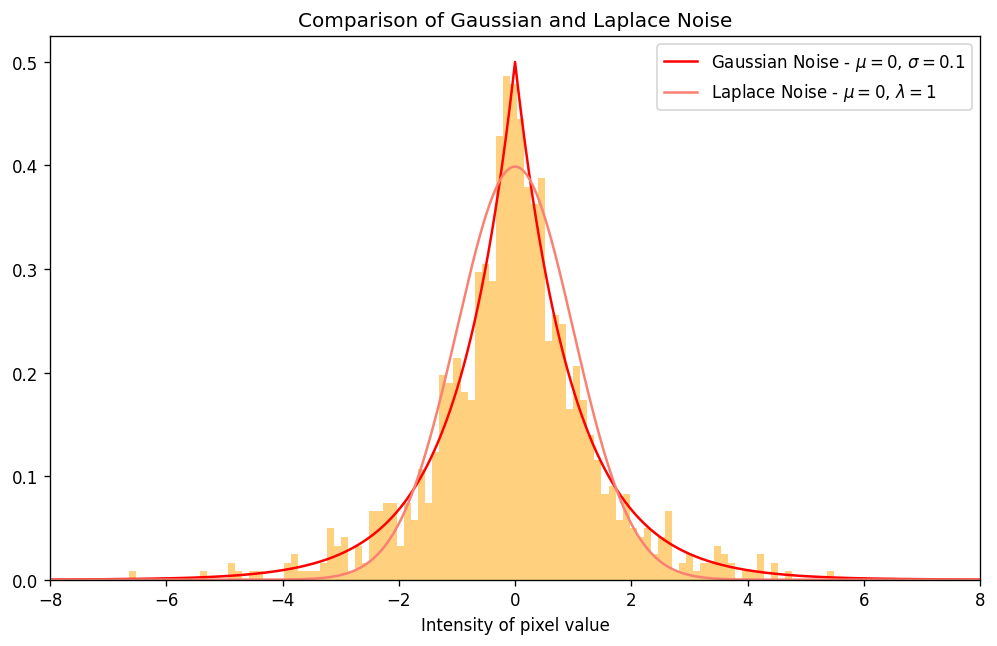}
\centering
\caption{PDF of Gaussian and Laplace noise}
\label{fig:gaus_lap_curve}
\end{figure}

The Gaussian Noise, also known as electronic noise because usually happens in amplifiers or detectors, generates disturbs in the gray values in the digital images. It can be described as below:

\begin{equation}
 \begin{aligned}
    P(g) = \sqrt{\frac{1}{2\pi \sigma^2}}^{-\frac{(g - \mu)^2}{2 \sigma ^2}}
 \end{aligned}
\end{equation}

where $g$ indicates the gray value, $\sigma$ the standard deviation and $\mu$ the mean. The Figure \ref{fig:gaus_lap_curve} shows the Gaussian Noise applied to an image extracted from the ORL dataset. Due to the properties of the normalized Gaussian noise curve, showed in the Figure \ref{fig:gaus_lap_curve}, the $70\%$ to $90\%$ of the noisy pixel values are between $\mu - \sigma$ and $\mu + \sigma$, considering $\mu=0$, $\sigma=0.1$ and 256 gray levels ($g$).

\begin{figure}[h]
\centering
\includegraphics[width=5.5
cm]{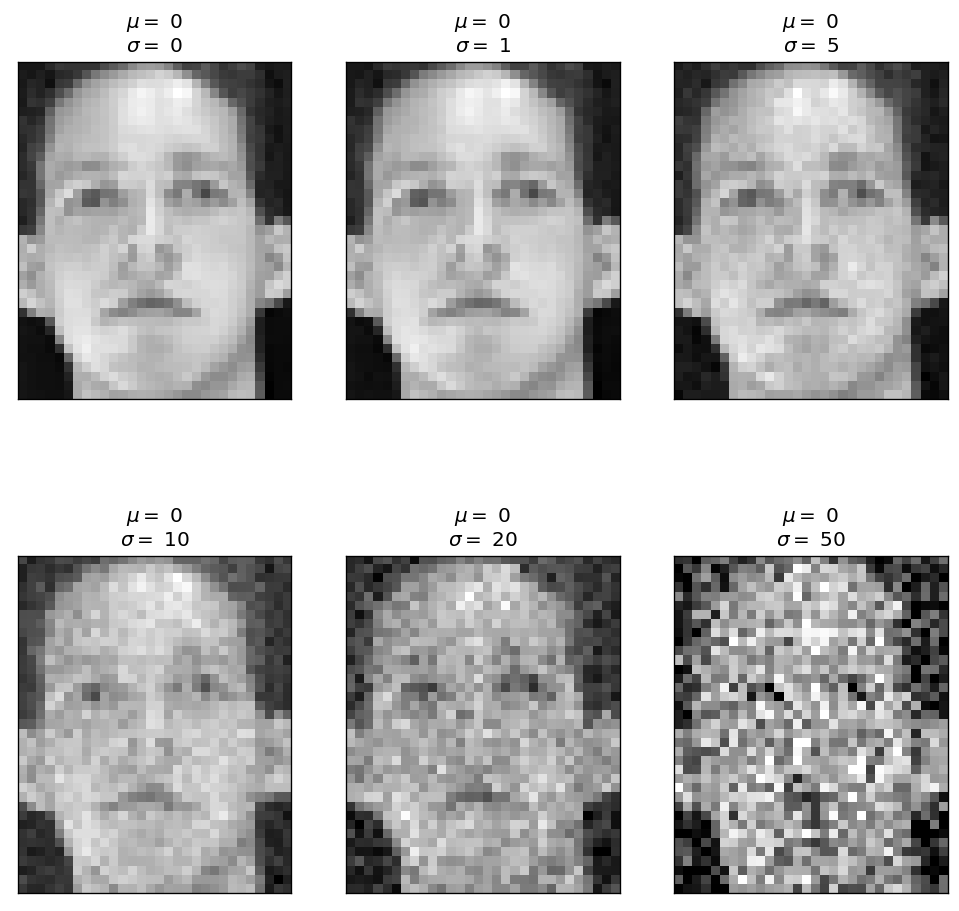}
\centering
\caption{Image with Gaussian noise with different values of $\sigma$.}
\label{fig:gaus_example}
\end{figure}

\subsubsection{Laplace Noise}
\begin{figure}[h]
\centering
\includegraphics[width=5.5cm]{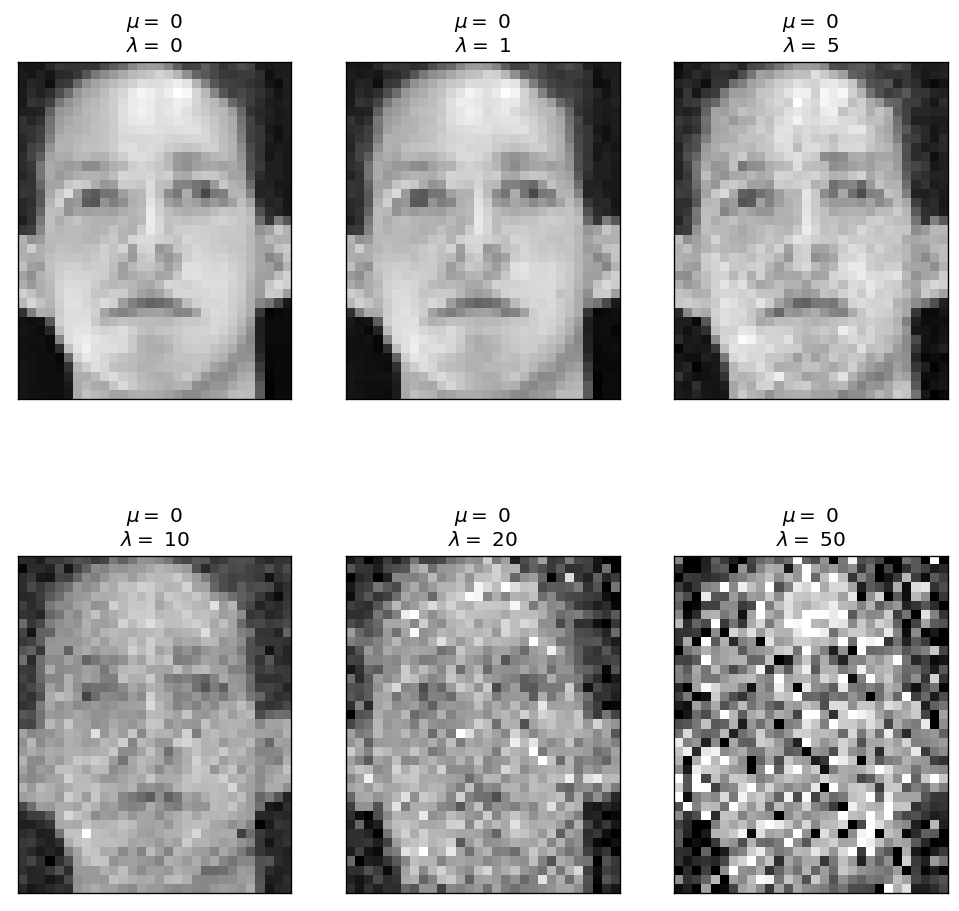}
\centering
\caption{Image with Laplace noise with different values of $\lambda$}
\label{fig:lap_example}
\end{figure}

Laplace Noise is generated by the Laplace Distribution and it is pretty similar to the Gaussian Distribution although it has a sharper peak around the mean value. The Figure \ref{fig:gaus_lap_curve} shows the comparison between the Gaussian and the Laplace Distribution, which can be described as below:

\begin{equation}
 \begin{aligned}
    f(x|\mu, \lambda) = \frac{1}{2\lambda}\exp(-\frac{|x - \mu|}{\lambda}) = \frac{1}{2\lambda} \begin{cases}
      \exp(-\frac{\mu - x}{\lambda}) & \text{if $x < \mu$}\\
      \exp(-\frac{x - \mu}{\lambda}) & \text{if $x \geq \mu$}
    \end{cases}    
 \end{aligned}
\end{equation}

where $\mu$ denotes the mean value and $\lambda$ is a scale parameter. As we can observe, the probability density function is expressed in terms of the absolute difference from the mean instead of the normal distribution, which is expressed in terms of squared difference from the mean. Consequently, the Laplace distribution has fatter tails than the normal distribution. In the Figure \ref{fig:lap_example}, can be seen the influence of the parameter $\lambda$ in an image extracted from the ORL dataset.

\section{Experiment and Results}

In this study, we create a series of experiments to prove the implementation of the Non-negative Matrix Factorization algorithms mentioned previously. Additionally, we compare the performance of these NMF algorithms to analyze their behavior in the presence of noise.

The datasets used to conduct the experiments are ORL and YaleB. The ORL dataset contains 400 images of 40 distinct subjects. The images were taken varying the lighting, facial expressions and facial details (some subjects wear glasses). We applied a $\text{reduce constant}=3$ to load the sample images into vectors with 1100 elements and resized to 92x112 pixels.

On the other hand, YaleB dataset contains 2414 images of 38 subjects under 9 different poses and 64 illumination conditions. We load the sample images,  using $\text{reduce constant}=3$, into vectors with 3584 elements.

\subsection{Metrics}
To compare the performance and robustness of the different NMF algorithms, we implement and use three evaluation metrics:

\begin{itemize}
  \item \textbf{Relative Reconstruction Errors (RRE):} Describes the similarity between the reconstructed matrix and the original cleaned matrix. Let $X$ denote the contaminated data, and $\widehat{X}$ denote the clean data. Let $U$ and $V$ denote the factorization results on $X$, the relative reconstruction errors can be defined as follows:
  
  \begin{equation}
    RRE=\frac{\|\widehat{X} - UV\|_F}{\|\widehat{X}\|_F}
  \end{equation}
  
  \item \textbf{Average Accuracy:} As the images in the dataset contain different subjects $n_p$, we perform a clustering, using K-means, on the reconstructed matrix with the num\_clusters equal to $n_p$ to get the prediction. Then, the average accuracy can be described as below:
  
  \begin{equation}
    ACC(Y, Y_{pred})=\frac{1}{n}\sum\limits_{i=1}^n1\{Y_{pred}(i) == Y(i)\}
  \end{equation}
  
  \item \textbf{Normalized Mutual Information (NMI):} Can be defined by:
  
  \begin{equation}
    NMI(Y, Y_{pred})=\frac{2 \ast I(Y, Y_{pred})}{H(Y) + H(Y_{pred})}
  \end{equation}
  where $I(\cdot, \cdot)$ is the mutual information and $H(\cdot)$ is entropy.
\end{itemize}

\subsection{Results}

Our experiments centered around two main themes: feature selection, and robustness to noise. Practical considerations such as run time performance and computational complexity played a vital role in how we set out to design our experiments, as our intention was to reflect real-world constraints. The first set of experiments relate to feature selection, looking at the sensitivity of robust NMF to the number of latent factors. We applied these findings to a number of simulated noise-based scenarios.

\subsubsection{Feature Selection}


We analyze the influence of the number of components or rank $k$ on the performance of NMF algorithms. We based this experiment on the ORL dataset. We added noise to the images to generate a set of polluted images $X$ and we trained the three NMF algorithms mentioned previously on this set of images. The dataset was contaminated with Gaussian Noise with a strength factor of $\sigma=0.05$. A range of number of components values was defined from 10 to 140 with a step of 10 to train the NMF algorithms. In addition, the parameter of regularization for L\textsubscript{1} Norm Robust NMF algorithm is set to $\lambda=0.04$. In this work the influence of the parameter $\lambda$ on the performance of the L\textsubscript{1} Norm Robust NMF algorithm is out of scope. However, the selection of the value of $\lambda$ is based on \cite{l1reg}, which illustrates that larger values of $\lambda$ indicates that the detected noise is more sparse and often this leads to higher precision and lower recall.\\

\begin{figure}[h]
\centering
\includegraphics[width=12cm]{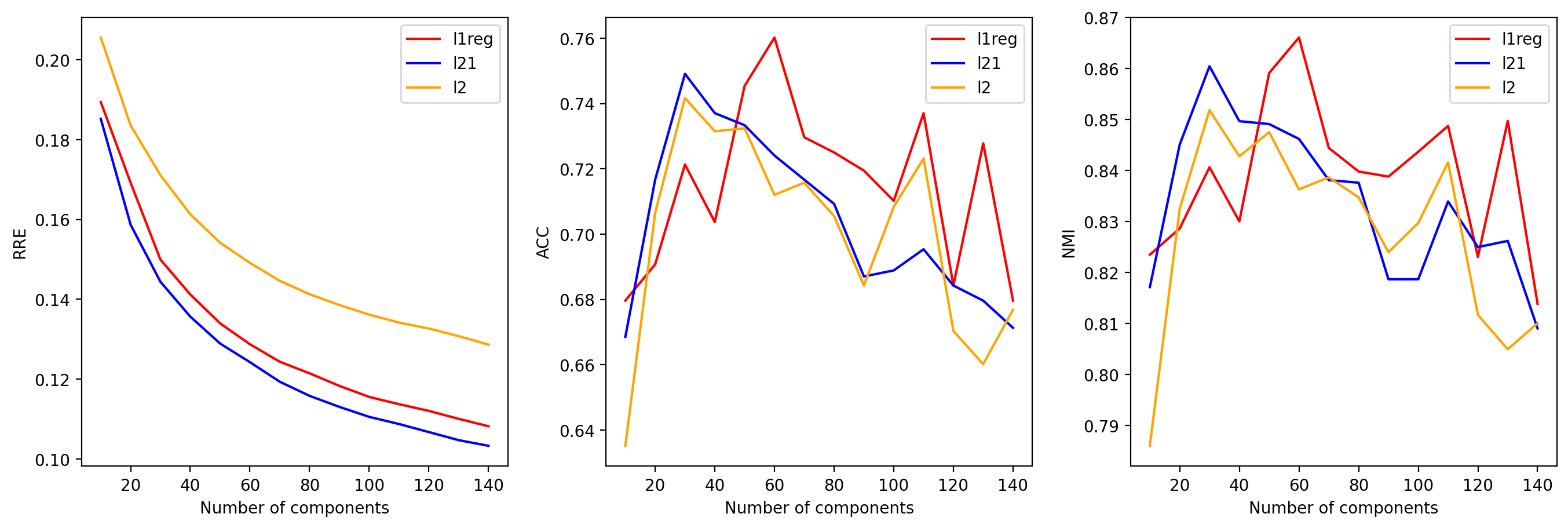}
\centering
\caption{Influence of number of components on RRE, ACC and NMI}
\label{fig:inf_ncomp}
\end{figure}

As we can observe in the Figure \ref{fig:inf_ncomp}, the left-hand panel shows that the Relative Reconstruction Errors (RRE) decreases as the number of components $k$ increases, which is an expected result. The more we increase the number of components, the more information is retained. Therefore, the reconstructed image given by the factorization results $U$ and $V$ is more similar to the original image.

On the other hand, the middle and right figures investigate the influence between the rank $k$ and the Average Accuracy (ACC) and Normalized Mutual Information (NMI). Nevertheless, it seems complicated to extract a correlation between the number of components and these metrics. It can be seen that the three algorithms have a similar performance. They achieve a maximum of around k=40-60 and then the ACC and NMI slightly decreases. However, as we mentioned before, the correlation between the $k$ and ACC and NMI is not easily deducible and a reason for this could be that the resolution on ORL dataset is too low.

\begin{figure}[h]
\centering
\includegraphics[width=12cm]{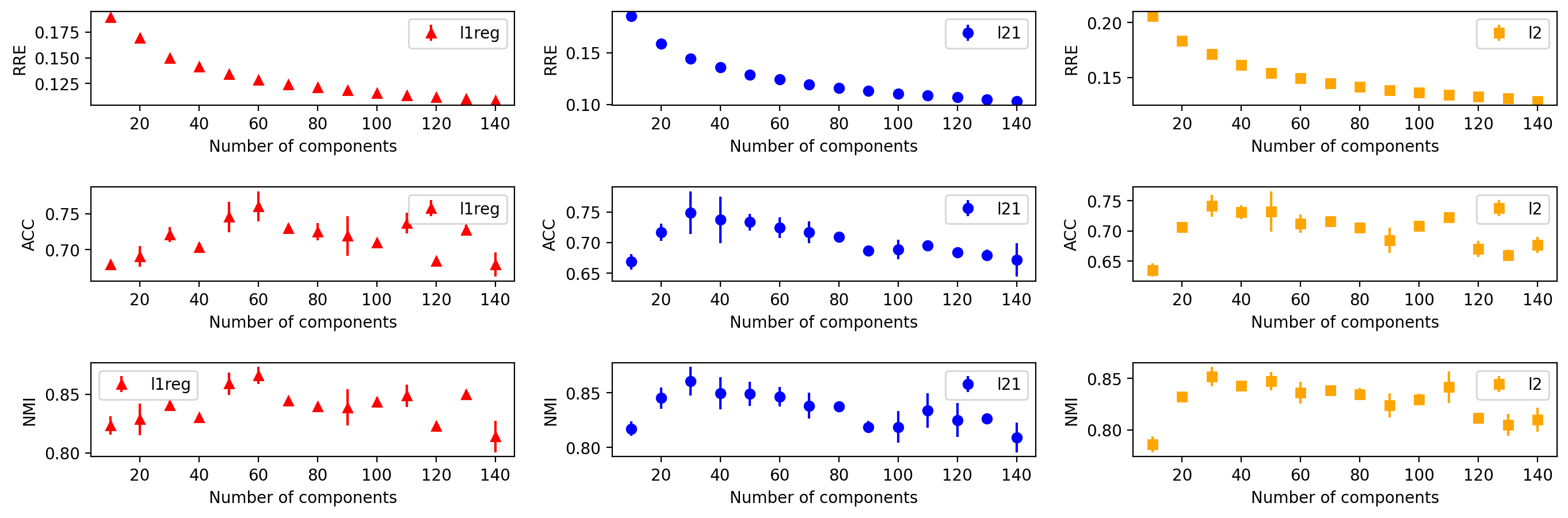}
\centering
\caption{Mean and standard deviation}
\label{fig:std_ncomp}
\end{figure}

To provide an estimate of the performance of the algorithms, each experiment was executed three times and the average and standard deviation were calculated for each measure. The Figure \ref{fig:std_ncomp} contains the mean value and standard deviation for each metric and algorithm in each experiment. It can be observed that the behavior of the three algorithms is approximately similar in each experiment. In addition, the standard deviation is generally small thus, the algorithms tend to be stable and to have the same performance in each iteration.

In general, as the number of components $k$ increases, the performance of the NMF algorithms also increase. Nonetheless, in dimensionality reduction the compression ratio usually decreases when the number of components rises thus, trade-off arises between the value of $k$ and the compression ratio.


\subsubsection{Robustness to Noise}
We applied our three additive-noise schemes to simulate a range of real-world scenarios; Salt \& Pepper, Laplacian, and Gaussian. Based on our feature selection analysis, each scenario used 60 components. Our methodology was adjusted between the two datasets to account for the computational complexity see Table \ref{table:exp-params}. The average and standard deviation for each experiment was calculated for each measure. The $L_{1}$ regularizer was set at 0.04 across all experiments.

\begin{table}[h!]
\centering
\begin{tabular}{||c||c c c || c c c||}
\hline
    \multicolumn{1}{||c||}{} & \multicolumn{3}{|c||}{ORL} & \multicolumn{3}{|c||}{YaleB}\\
\hline
 Model Param. & Max Iter. & No. Exp. & Repeated & Max Iter. & No. Exp. & Repeated \\ [1ex] 
 \hline\hline
 $L_{1}$-norm & 300 & 3 & 3 & 100 & 3 & 2\\
 $L_{2,1}$-norm & 300 & 3 & 3 & 100 & 3 & 2\\ 
 $L_{2}$-norm & 300 & 3 & 3 & 100 & 3 & 2\\
 \hline\hline
\end{tabular}
\caption{Table of parameters used in the experiments}
\label{table:exp-params}
\end{table}

It is considered that most of the communication and computer systems may be affected by Gaussian noise specifically.The noise may come from different natural sources. Therefore, we interested in measuring the effect of the Gaussian noise scale on the provided datasets. We performed experiments on different sigma values on the added Gaussian noise in order to obtain better insight on how the parameter affects our evaluation metrics such as RRE, ACC and NMI.

\begin{figure}[h]
\centering
\includegraphics[width=12cm]{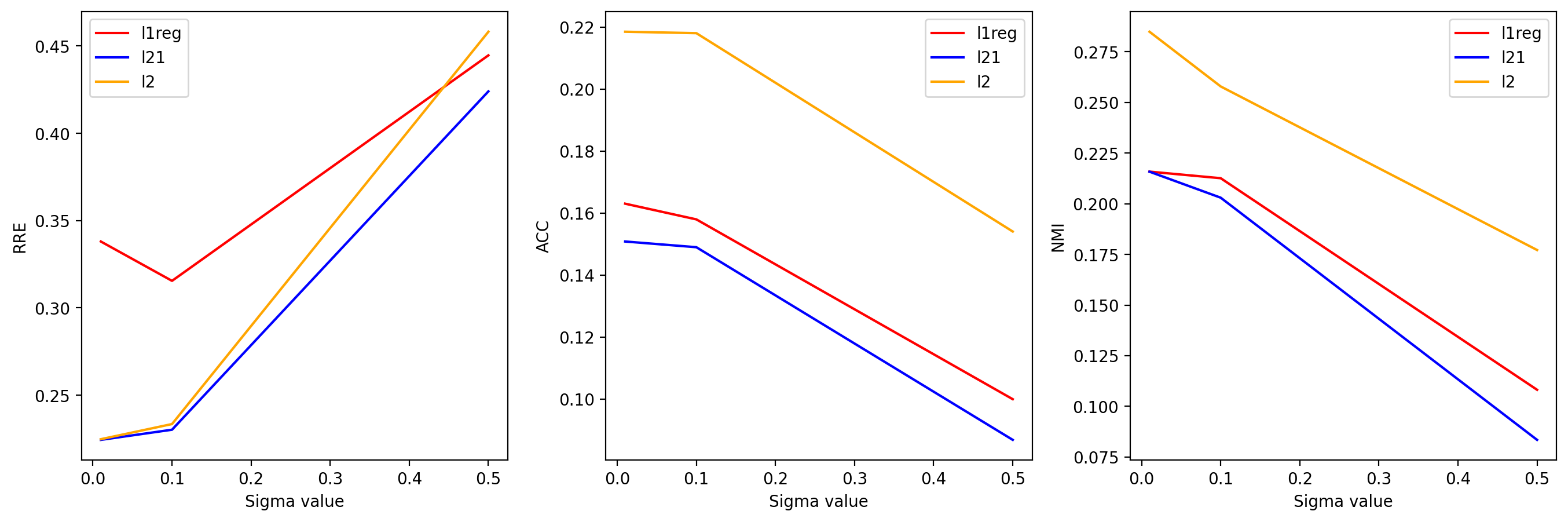}
\centering
\caption{Influence of Sigma on REE, ACC and NMI}
\label{figure:Sigmamean}
\end{figure}


We explored different values of sigma, specifically 0.01, 0.1 and 0.5 for our three implemented algorithms. Figure \ref{figure:Sigmamean} indicates the mean results of the experiments on the evaluation metrics.

Gaussian noise is defined by its mean as well as its standard deviation to the mean Sigma. This value may be interpreted physically as related to the noise power. The higher this value goes, the stronger the noise in the images. Therefore, it can be observed from figure \ref{figure:Sigmamean} that with the sigma  value of 0.001, all three algorithm performs significantly better. It can also be observed the L2 achieved higher result in all three evaluation metrics at every sigma value. L2 NMF method with NNDSVD initiation also achieved the best result at sigma value of 0.01.

\begin{figure}[h]
\centering
\includegraphics[width=12cm]{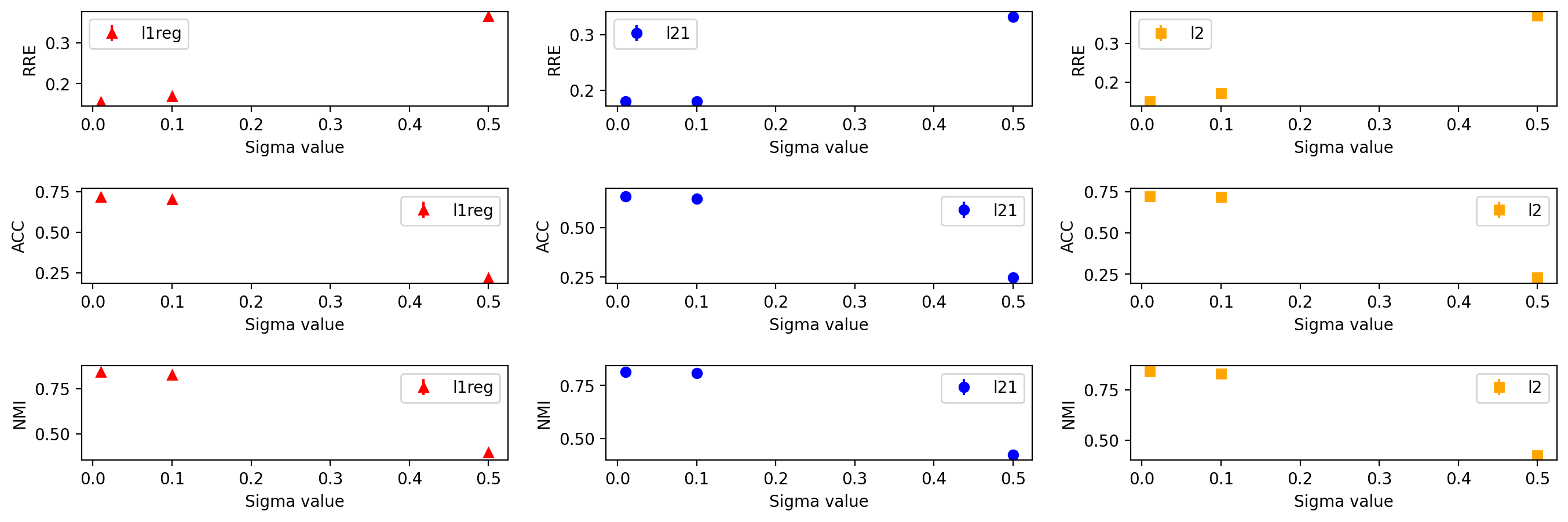}
\centering
\caption{Mean and standard deviation}
\label{figure:Stdmamean}
\end{figure}

It can be easily observed that there is no significant amount of standard deviation at any sigma value for almost all metrics. Therefore, picking 0.01 as the the optimal sigma value for the Gaussian noise was our optimal decision.

\begin{figure}[h]
\centering
\includegraphics[width=12cm]{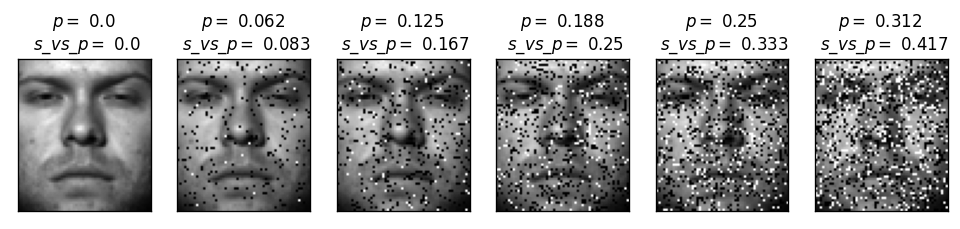}
\centering
\caption{Span of Salt \& Pepper noise.}
\label{fig:sp-noise}
\end{figure}

The Salt \& Pepper noise scenarios as illustrated in Figure \ref{fig:sp-noise} consisted of a spread of both the percentage $p$ and range of additive noise $svp$ designed to push the limits each technique. Each method suffered a precipitous decline in accuracy for proportions of noise above 40 percent. The $L_{2,1}$-norm performed only marginally better overall, but was able to consistently produce better results over $L_{1}$-norm and $L_{2}$-norm across RRE in particular. We observed the most substantial spread in ACC in Table \ref{table:results}. We might surmise $L_{2,1}$ is more susceptible to a larger range in white to black noise.

\begin{figure}[h]
\centering
\includegraphics[width=12cm]{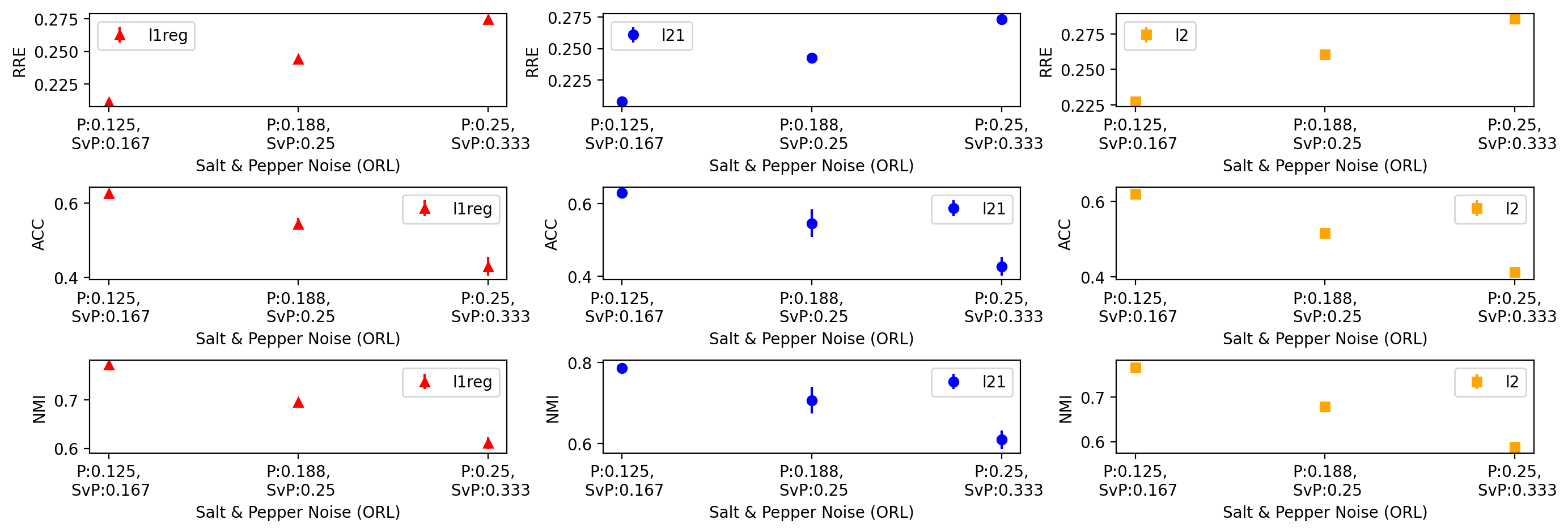}
\centering
\caption{Mean and standard deviation of ORL Salt \& Pepper experiments.}
\label{fig:salt-pepper-orl-stddev}
\end{figure}

Figure \ref{fig:salt-pepper-orl-stddev} shows a strong convergence on our ORL experiments for all three methods, where as at the same time might also suggest $L_{1}$-norm is adversely impacted by a lower number of iterations on the more complex YaleB domain.

\begin{figure}[h]
\centering
\includegraphics[width=12cm]{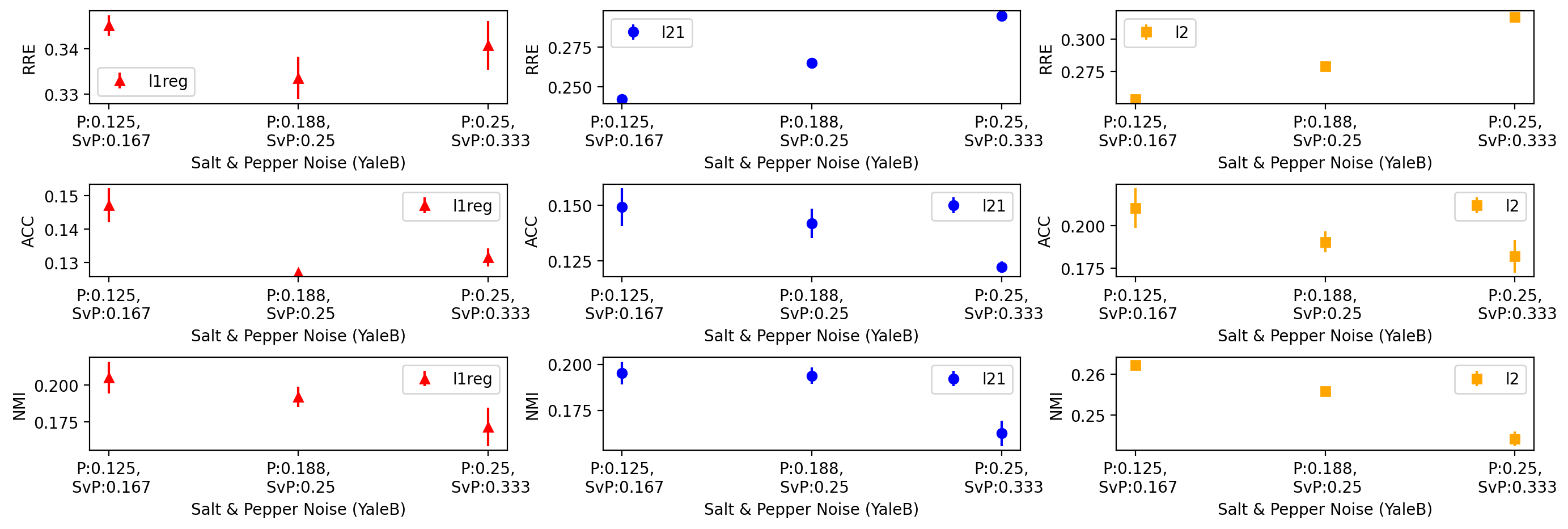}
\centering
\caption{Mean and standard deviation of YaleB Salt \& Pepper experiments.}
\label{fig:salt-pepper-yaleb-stddev}
\end{figure}

We set out to identify the sensitivity of robust methods to a laplacian distribution using a sliding scale factor, increasing linearly from 0.06. We hypothesized a strong performance in both robust methods, but set out to identify the limits of $L_{2,1}$. We can see $L_{1}$-norm gradually outperform both $L_{2,1}$-norm and $L_2$-norm in Table \ref{table:results} as a larger range of noise is introduced. 

Analysis of Figure \ref{fig:laplace-yaleb-stddev} is consistent with what we found in our Salt \& Pepper experiments. The slower convergence of $L_{1}$-norm makes its performance susceptible to lower iterations. The computationally more complex nature of $L_{1}$-norm made increasing the number of iterations impractical. By contrast, the relatively more stable $L_{2,1}$ failed to out-perform $L_{2}$ in our Laplacian experiments as illustrated in Figure \ref{fig:laplace-yaleb-stddev}


\begin{figure}[h]
\centering
\includegraphics[width=12cm]{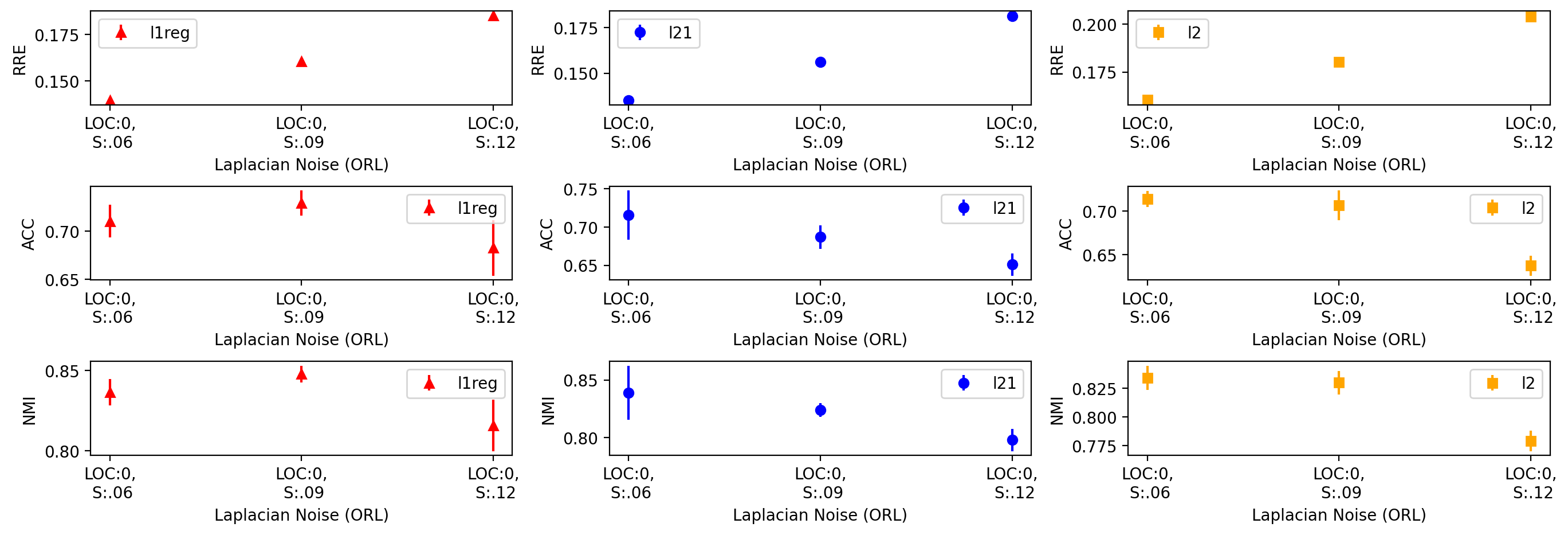}
\centering
\caption{Mean and standard deviation of ORL Laplacian experiments.}
\label{fig:laplace-orl-stddev}
\end{figure}

\begin{figure}[h]
\centering
\includegraphics[width=12cm]{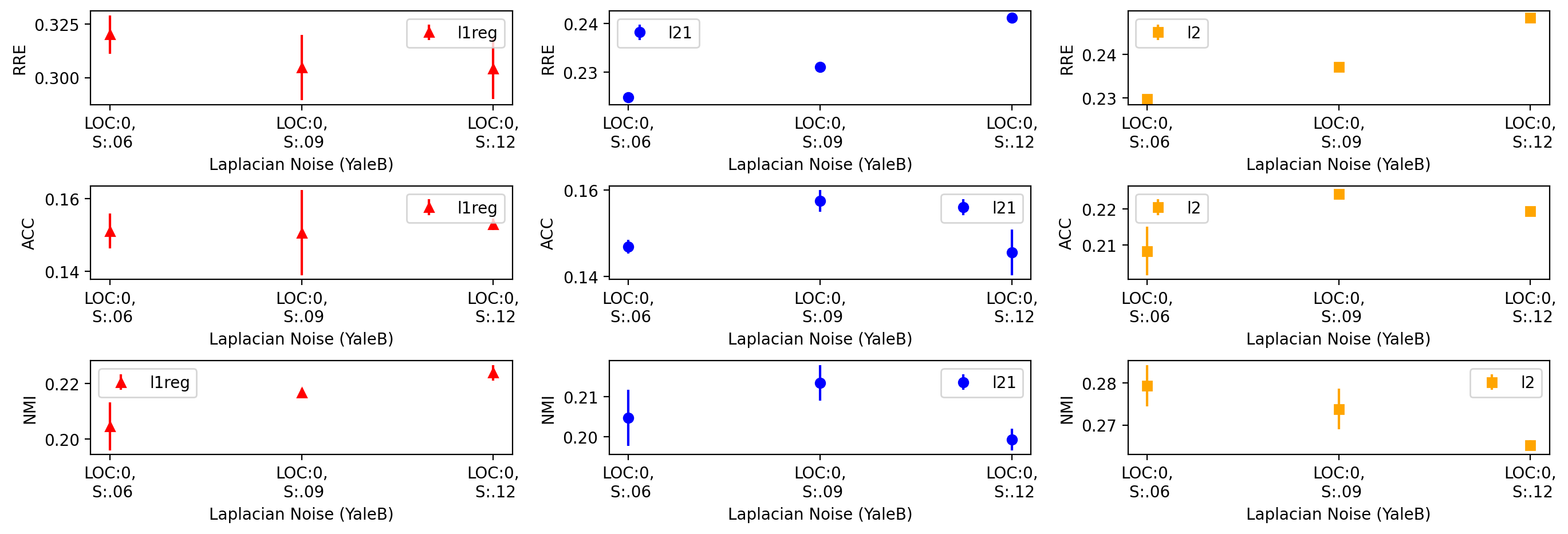}
\centering
\caption{Mean and standard deviation of YaleB Laplacian experiments.}
\label{fig:laplace-yaleb-stddev}
\end{figure}

\pagebreak

\begin{table}[h]
\centering
\begin{tabular}{||c||c c c || c c c||}

\hline
    \multicolumn{1}{||c||}{Experiment} & \multicolumn{3}{|c||}{ORL} & \multicolumn{3}{|c||}{YaleB}\\
\hline
 \hline\hline
 noise\_type=S\&P (p=0.125, svp=0.167) & RRE & ACC & NMI & RRE & ACC & NMI \\ [1ex] 
 \hline\hline
 $L_{1}$-norm robust & 0.211 & 0.627 & 0.772 & 0.345 & 0.147 & 0.205\\
 $L_{2,1}$-norm & \textbf{0.208} & \textbf{0.631} & \textbf{0.787} & \textbf{0.242} & 0.149 & 0.195\\ 
 $L_{2}$-norm & 0.227 & 0.619 & 0.767 & 0.254 & \textbf{0.21} & \textbf{0.262}\\
 \hline\hline
 noise\_type=S\&P (p=0.188, svp=0.25) & RRE & ACC & NMI & RRE & ACC & NMI \\ [1ex] 
 \hline\hline
 $L_{1}$-norm robust & 0.244 & 0.544 & 0.695 & 0.334 & 0.127 & 0.192\\
 $L_{2,1}$-norm & \textbf{0.243} & \textbf{0.546} & \textbf{0.707} & \textbf{0.265} & 0.142 & 0.194\\ 
 $L_{2}$-norm & 0.261 & 0.516 & 0.678 & 0.279 & \textbf{0.191} & \textbf{0.256}\\
 \hline\hline
 noise\_type=S\&P (p=0.25, svp=0.333) & RRE & ACC & NMI & RRE & ACC & NMI \\ [1ex] 
 \hline\hline
 $L_{1}$-norm robust & 0.275 & \textbf{0.43} & \textbf{0.611} & 0.341 & 0.132 & 0.171\\ 
 $L_{2,1}$-norm & \textbf{0.273} & 0.428 & 0.61 & \textbf{0.294} & 0.122 & 0.162\\ 
 $L_{2}$-norm & 0.286 & 0.413 & 0.588 & 0.317 & \textbf{0.182} & \textbf{0.244}\\
 \hline\hline
 noise\_type=laplace (loc=0, scale=0.06) & RRE & ACC & NMI & RRE & ACC & NMI \\ [1ex] 
 \hline\hline
 $L_{1}$-norm robust & 0.139 & 0.71 & 0.836 & 0.32 & 0.151 & 0.205 \\
 $L_{2,1}$-norm & \textbf{0.135} & \textbf{0.716} & \textbf{0.839} & \textbf{0.225} & 0.147 & 0.205\\ 
 $L_{2}$-norm & 0.161 & 0.714 & 0.834 & 0.23 & \textbf{0.208} & \textbf{0.279}\\
 \hline\hline
noise\_type=laplace (loc=0, scale=0.09) & RRE & ACC & NMI & RRE & ACC & NMI \\ [1ex] 
 \hline\hline
 $L_{1}$-norm robust & 0.161 & \textbf{0.729} & \textbf{0.848} & 0.305 & 0.151 & 0.217 \\
 $L_{2,1}$-norm & \textbf{0.156} & 0.687 & 0.824 & \textbf{0.231} & 0.157 & 0.213\\ 
 $L_{2}$-norm & 0.18 & 0.706 & 0.83 & 0.237 & \textbf{0.224} & \textbf{0.274}\\
 \hline\hline
noise\_type=laplace (loc=0, scale=0.12) & RRE & ACC & NMI & RRE & ACC & NMI \\ [1ex] 
 \hline\hline
 $L_{1}$-norm robust & 0.185 & \textbf{0.682} & \textbf{0.816} & 0.304 & 0.153 & 0.224 \\
 $L_{2,1}$-norm & \textbf{0.181} & 0.651 & 0.798 & \textbf{0.241} & 0.146 & 0.199\\ 
 $L_{2}$-norm & 0.204 & 0.637 & 0.779 & 0.248 & \textbf{0.219} & \textbf{0.265}\\
 \hline\hline
noise\_type=gaussian (mean=0, sigma=0.01) & RRE & ACC & NMI & RRE & ACC & NMI \\ [1ex] 
 \hline\hline
 $L_{1}$-norm robust & 0.123 & 0.731 & 0.85 & 0.338 & 0.163 & 0.216\\
 $L_{2,1}$-norm & \textbf{0.119} & \textbf{0.743} & \textbf{0.858} & \textbf{0.224} & 0.151 & 0.216\\ 
 $L_{2}$-norm & 0.143 & 0.73 & 0.847 & 0.225 & \textbf{0.218} & \textbf{0.285}\\
 \hline\hline
noise\_type=gaussian (mean=0, sigma=0.1) & RRE & ACC & NMI & RRE & ACC & NMI \\ [1ex] 
 \hline\hline
 $L_{1}$-norm robust & 0.148 & 0.717 & 0.83 & 0.316 & 0.158 & 0.213 \\
 $L_{2,1}$-norm & \textbf{0.143} & \textbf{0.719} & \textbf{0.834} & \textbf{0.23} & 0.149 & 0.203\\ 
 $L_{2}$-norm & 0.167 & 0.701 & 0.823 & 0.233 & \textbf{0.218} & \textbf{0.258}\\
 \hline\hline
    noise\_type=gaussian (mean=0, sigma=0.5) & RRE & ACC & NMI & RRE & ACC & NMI \\ [1ex] 
 \hline\hline
 $L_{1}$-norm robust & 0.38 & 0.247 & \textbf{0.426} & 0.445 & 0.1 & 0.108 \\
 $L_{2,1}$-norm & \textbf{0.379} & \textbf{0.252} & 0.419 & \textbf{0.424} & 0.087 & 0.083\\ 
 $L_{2}$-norm & 0.375 & 0.238 & 0.422 & 0.458 & \textbf{0.154} & \textbf{0.177}\\
 \hline\hline
\end{tabular}
\caption{Table of results}
\label{table:results}
\end{table}

\pagebreak

\section{Conclusion}

In this project we presented and implemented the formulation of three different Non-negative Matrix Factorization algorithms. We defined several types of noise to contaminate the samples images of the ORL and YaleB datasets and we analyzed the performance of these algorithms for different magnitude of noise.

It could be observed that as the number of components $k$ increases the Relative Reconstruction Errors of the NMF algorithms decreases. However, it was complicated to extract a correlation for the Average Accuracy and Normalized Mutual Information metrics and we concluded that this can be generated for the low resolution of the ORL dataset.

On the other hand, we analyzed the robustness to noise of each algorithm. We observed that the more we increase the noise scale, the more RRE is increased. All three algorithms performs significantly better for low values of noise, particularly the L\textsubscript{2-1}-norm NMF algorithm which achieved the best result with Gaussian Noise. In parallel, as we expected, when the effect of the noise scale is increased, the ACC and NMI values decreases. Additionally, when the sample images were contaminated with Salt and Pepper Noise the algorithms suffered a precipitous decline in accuracy for proportions of noise above 40 percent. However, we found that the performance of L\textsubscript{1} robust norm and L\textsubscript{2-1}-norm were pretty similar although the L\textsubscript{2-1} produced better results. 

Finally, we found that the computationally cost of L\textsubscript{1}-norm algorithm made increasing the number of iterations impractical and the execution time was considerably higher than for the other algorithms.

In the future, we can extend this study in the following aspects:

\begin{itemize}
    \item Influence of the regularization parameter $\lambda$ on the performance of L\textsubscript{1} Norm Robust NMF algorithm \cite{l1reg}
    \item Influence between the rank $k$ and the Average Accuracy and Normalized Mutual Information on YaleB dataset. Analyze if the low resolution on ORL dataset explain the low correlation between the number of components and ACC and NMI.
    \item Extend the NNDSVD initialization method to the other NMF algorithms.
\end{itemize}

Common themes and considerations emerged from our research of weight-based robust NMF algorithms such as $L_1$, and $L_{2,1}$ norm. Reflecting on this work, we identified an opportunity to further explore the sensitivity of these robust NMF techniques to the initial state of the factor matrices.

\clearpage
\vskip 1in

\printbibliography

\clearpage
\section*{Appendix}

\href{https://github.com/alejandrods/Analysis-of-the-robustness-of-NMF-algorithms}{\textit{Click here to access to the repository with the code.}}{}\\

The code of this work is provided in the Jupyter Notebook called \emph{Analysis\_of\_the\_robustness\_of\_NMF\_algorithms.ipynb} and the experiments were executed using Python 3x in Google Colab. 

\begin{figure}[h]
\centering
\includegraphics[width=12cm]{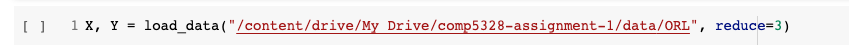}
\centering
\caption{Provide a path to the target dataset.}
\label{fig:drive_1}
\end{figure}

To execute the code you just need to go through the notebook and executing all cells. The first section called \emph{Load and preprocessing data} contains the function to load and preprocess the images. In this section, you should add the path to the folder with the datasets, in case you use Google Colab you need to add the datasets to your Drive account and mount your Drive folder into the notebook (figure \ref{fig:drive_1}).

The section \emph{Utils Functions} contains functions that are used in the implementation of our algorithms. The functions to generate noise are contained in this section as well as the functions to calculate the metrics. Furthermore, the section called \emph{Non-negative Matrix Factorization} contains the implementation of our algorithms and we used the ORL dataset to test them. Finally, the section \emph{Experiments} has the experiments we have executed to analyze the performance of the NMF algorithms. Additionally, the section \emph{Appendix} contains some images generated for the report.

\end{document}